\documentclass[letterpaper, 10 pt, conference]{ieeeconf}  

\usepackage{amssymb}
\usepackage{pifont}

\usepackage{cite}
\usepackage{amsmath,bm}
\usepackage[T1]{fontenc}    
\usepackage{hyperref}       
\usepackage{url}            
\usepackage{booktabs}       
\usepackage{multirow}
\usepackage{amsfonts}       
\usepackage{nicefrac}       
\usepackage{microtype}      
\usepackage{subcaption}
\usepackage{caption}
\usepackage{graphicx}
\usepackage[ruled,vlined]{algorithm2e}
\usepackage{MnSymbol}
\usepackage{dsfont}

\usepackage{cleveref}

\usepackage{amsmath}
\usepackage{amssymb}
\usepackage{mathtools}
\usepackage{bm}
\usepackage{bbm}
\usepackage{comment}
\usepackage{makecell}
\usepackage{xcolor}

\usepackage{amssymb}
\usepackage{pifont}

\usepackage{comment}

\IEEEoverridecommandlockouts                              

\title{\LARGE \bf
Beyond Domain Randomization: Event-Inspired Perception \\ for Visually Robust Adversarial Imitation from Videos}

\author{Andrea Ramazzina$^{*1,2}$, Vittorio Giammarino$^{*3}$, Matteo El-Hariry$^{4}$ and Mario Bijelic$^{5,6}$
\thanks{*Denotes equal contribution}%
\thanks{$^{1}$ Mercedes-Benz AG, Germany, $^{2}$ Technical University of Munich, Germany, $^{3}$ Purdue University, USA, $^{4}$ Space Robotics Research Group, SnT, University of Luxembourg, $^{5}$ Torc Robotics, USA, $^{6}$ Princeton University, USA, $^{7}$ https://github.com/VittorioGiammarino/Eb-LAIfO}}

\begin{document}

\maketitle
\thispagestyle{empty}
\pagestyle{empty}


\begin{abstract}
Imitation from videos often fails when expert demonstrations and learner environments exhibit domain shifts, such as discrepancies in lighting, color, or texture. While visual randomization partially addresses this problem by augmenting training data, it remains computationally intensive and inherently reactive, struggling with unseen scenarios. We propose a different approach: instead of randomizing appearances, we eliminate their influence entirely by rethinking the sensory representation itself. Inspired by biological vision systems that prioritize temporal transients (e.g., retinal ganglion cells) and by recent sensor advancements, we introduce event-inspired perception for visually robust imitation. Our method converts standard RGB videos into a sparse, event-based representation that encodes temporal intensity gradients, discarding static appearance features. This biologically grounded approach disentangles motion dynamics from visual style, enabling robust visual imitation from observations even in the presence of visual mismatches between expert and agent environments. By training policies on event streams, we achieve invariance to appearance-based distractors without requiring computationally expensive and environment-specific data augmentation techniques. Experiments across the DeepMind Control Suite  and the Adroit platform for dynamic dexterous manipulation show the efficacy of our method. Our code is publicly available at \href{https://github.com/VittorioGiammarino/Eb-LAIfO}{Eb-LAIfO}$^7$.
\end{abstract}
\section{INTRODUCTION}
\label{sec:intro}
Visual Imitation from Observations (V-IfO) has emerged as a rapidly growing research area in recent years \cite{liu2022visual, giammarino2023adversarial}. This approach presents a compelling proposition for the future of robotics, offering an efficient and scalable method for teaching new skills to autonomous systems, and a possible alternative to tedious and complex handcrafting of ad-hoc rewards.
While significant progress has been made in addressing the fundamental challenges of V-IfO—namely, the partial observability of decision-making processes and the lack of explicit expert action information—current solutions still face important limitations. Specifically, state-of-the-art (SOTA) end-to-end algorithms typically assume environmental consistency between expert demonstrations and agent deployment scenarios~\cite{liu2022visual, giammarino2023adversarial}. This assumption proves problematic in real-world applications, where environmental conditions can vary significantly.  

Recently, various methods~\cite{stadie2017third, okumura2020domain, cetin2020domain, choi2023domain, giammarino2024visually} have been proposed to address this gap; with the most effective approach~\cite{giammarino2024visually} relying on contrastive learning and data augmentation for invariant feature extraction. This reliance limits the applicability of this approach, as visual data augmentation techniques may be unavailable in certain domains or computationally prohibitive. Additionally, designing effective augmentations requires prior knowledge of the expert domain or assumptions on the types of mismatches, further restricting its usability.

We depart from this line of work and propose to eliminate the visual domain gap at its origin, namely the input signal representation. 
Specifically, we draw inspiration from biological vision systems—where retinal ganglion cells encode temporal intensity changes \cite{kolb1995organization} rather than absolute luminance—and from the recent advancements in event cameras and their underlying image formation model. Building on these insights, we redefine the agent’s perceptual interface to focus on discretized temporal gradients (edges, motion). Consequently, our method inherently discards appearance-based distractors (lighting, color, texture) while preserving the motion dynamics features critical for control.
We evaluate this technique, integrated with a SOTA end-to-end V-IfO algorithm, on two challenging benchmarks: the DeepMind Control Suite~\cite{tassa2018deepmind} for locomotion under visual perturbations (e.g., color randomization, lighting change) and the Adroit platform for robotic dexterous manipulation~\cite{Kumar2016thesis}, requiring fine-grained motion alignment.
Specifically, we make the following contributions:
\begin{itemize}
  \item We design a lightweight event-inspired perception module, eliminating appearance-based distractors (e.g., color, texture, lighting) while preserving motion-critical temporal gradients.
  \item Our method synthesizes event streams directly from RGB videos, bypassing computationally expensive domain randomization or handcrafted augmentation pipelines. 
  \item By demonstrating robust imitation from synthetic event streams, we unlock the potential for real-world deployment with event cameras—low-power, high-temporal-resolution sensors that natively capture the temporal dynamics our method exploits.
\end{itemize}

\section{RELATED WORK}
\label{sec:related_work}

\paragraph{Imitation from expert videos} The Imitation Learning (IL) paradigm involves training agents to replicate expert behavior using task demonstrations, typically represented as state-action pairs. Among IL methods, Adversarial Imitation Learning (AIL) \cite{ho2016generative, fu2017learning} stands out as a flexible and effective approach, modeling IL as an adversarial interaction between a discriminator and the agent’s policy, where the discriminator distinguishes whether a state-action pair originates from the expert's or the agent's behavior policy. Building on principles from inverse Reinforcement Learning (RL) \cite{russell1998learning, ng2000algorithms, abbeel2004apprenticeship, ziebart2008maximum}, AIL uses the discriminator’s output as a reward signal to guide the agent’s training through RL. Recent adaptations of AIL extend its applicability to partially observable environments, addressing missing information \cite{gangwani2020learning}, and to visual IL, where agents learn directly from video frames instead of structured states \cite{rafailov2021visual}. A key variation, Imitation from Observation~\cite{torabi2018generative, yang2019imitation, cheng2021guaranteed}, eliminates the need for action labels, making it more practical but also more challenging, as agents must infer expert behavior solely from state sequences. When these state sequences consist of video data without action labels, the problem is referred to as V-IfO. SOTA V-IfO methods include PatchAIL \cite{liu2022visual}, which applies AIL directly to pixel space using a PatchGAN discriminator \cite{isola2017image, zhu2017unpaired}, and LAIfO \cite{giammarino2023adversarial}, which learns a latent representation of agent states. However, these methods assume that the expert and the learner operate within the same decision-making framework, an assumption rarely met in real-world applications.

\paragraph{Imitation from videos with mismatches} Our research tackles visual imitation from observations under domain mismatches (V-IfO with mismatches), a problem also known as third-person IL \cite{stadie2017third}, domain-adaptive IL \cite{kim2020domain}, or cross-domain IL \cite{raychaudhuri2021cross}. Prior solutions fall into two categories: sequential and end-to-end approaches. Sequential methods break the problem into distinct learning stages, tackling them consecutively~\cite{liu2018imitation, sermanet2018time, smith2020avid, giammarino2023opportunities, zhang2023slomo}. In contrast, end-to-end methods~\cite{stadie2017third, okumura2020domain, cetin2020domain, choi2023domain, giammarino2024visually} aim to learn online, mapping pixel observations directly to actions without intermediate learning stages. In end-to-end methods, a common strategy for handling domain mismatches is to extract domain-invariant features. For instance, in \cite{stadie2017third}, adversarial learning is used to align feature distributions across domains, while DisentanGAIL \cite{cetin2020domain} enforces a mutual information constraint to preserve task-relevant information in the feature space. More recently, C-LAIfO \cite{giammarino2024visually} has leveraged contrastive learning for domain-invariant feature extraction, where these features are used across the entire AIL pipeline for both reward inference and policy learning. Unlike these methods, our approach avoids the need for domain-invariant feature learning. Instead, we disentangle motion dynamics from visual style at the observation level, effectively reducing the problem to standard V-IfO without mismatches. This eliminates the need for costly feature extraction steps while ensuring robust imitation performance. Other works address domain mismatches using generative approaches, such as learning domain-agnostic latent dynamics~\cite{okumura2020domain} or transforming expert videos to match the agent’s domain via cycle-consistent adversarial networks~\cite{choi2023domain}. However, these methods are computationally demanding and challenging to implement, as they rely on costly generative steps. In contrast, our approach filters out non-essential signals directly at the observation level, eliminating the need for additional learning steps. As a result, it is significantly more computationally efficient than learning-based alternatives.
\paragraph{Event-based cameras in robotics}
Unlike conventional cameras that capture complete images at fixed intervals, event cameras detect brightness changes asynchronously and independently at each pixel. Event cameras produce a variable-rate stream of digital "events," with each event signaling a specific threshold change in brightness at a particular pixel location and time~\cite{gallego2020event}.
Specifically, each pixel stores the log intensity value whenever it generates an event and continuously monitors for significant deviations from this stored value~\cite{gallego2020event}. 
Due to their high temporal resolution and low power characterisitics, event-based cameras have rapidly been adopted in vision-based robotics, enabling the development of highly responsive policies that directly convert sparse event data into control commands \cite{silva2025recurrent}.
These systems have been deployed across various robotic platforms with promising results. For example, previous works have successfully trained quadruped robots to intercept fast-moving objects by applying RL to event-based visual inputs \cite{forrai2023eventcatching}.
In aerial robotics, drones equipped with event cameras have demonstrated collision avoidance capabilities against dynamic obstacles using model-based algorithms \cite{bhattacharya2024monocular}, showcasing the exceptional temporal resolution of these sensors.
The applications of event-based vision extend beyond robotics into computer vision domains, including classical computer vision tasks such as video deblurring \cite{kim2024deblurring}.
Other lines of work have also developed multimodal approaches that integrate event-based and conventional frame-based vision, capitalizing on their complementary advantages. For instance, \cite{pellerito2024deep} proposed an end-to-end learned visual odometry framework that combines asynchronous event streams with standard image data, achieving improved efficiency and accuracy in challenging visual conditions.

While we do not directly use an event camera, we simulate its underlying functioning, extracting discrete events (in a synchronous manner) from a duplet of RGB frames, and use this representation as input for performing direct imitation from videos. 

\section{PRELIMINARIES}
\label{sec:preliminaries}

\paragraph{Modeling the visual mismatch in partially observable Markov decision process}  In the standard formulation of the V-IfO problem, both the expert and the agent are assumed to operate within the same Partially Observable Markov Decision Process (POMDP), described by the tuple $(\mathcal{S}, \mathcal{A}, \mathcal{X}, \mathcal{T}, \mathcal{U}, \mathcal{R}, \rho_0, \gamma)$. We define $\mathcal{S}$ as the state space, $\mathcal{A}$ the action space, $\mathcal{X}$ the observation space and $\mathcal{T}: \mathcal{S} \times \mathcal{A} \to P(\mathcal{S})$ the transition probability function where $P(\mathcal{S})$ denotes the set of probability distributions over $\mathcal{S}$. Furthermore, $\mathcal{U}: \mathcal{S} \to P(\mathcal{X})$ is the observation probability function, $\mathcal{R}: \mathcal{S} \times \mathcal{A} \to \mathbb{R}$ the reward function, $\rho_0 \in P(\mathcal{S})$ the initial state distribution, and $\gamma \in [0,1)$ the discount factor. In this work, we consider the expert and the agent interacting with different, but related, decision processes. Specifically, we define two distinct POMDPs: a \textit{source-POMDP} for the expert and a \textit{target-POMDP} for the agent. The target-POMDP is defined by $(\mathcal{S}, \mathcal{A}, \mathcal{X}, \mathcal{T}, \mathcal{U}_T, \mathcal{R}, \rho_0, \gamma)$, while the source-POMDP is characterized by $(\mathcal{S}, \mathcal{A}, \mathcal{X}, \mathcal{T}, \mathcal{U}_S, \mathcal{R}, \rho_0, \gamma)$. The key distinction between these two lies in their observation probability functions, $\mathcal{U}_S$ and $\mathcal{U}_T$, which govern the relationship between the environment state and the agent's or expert's observations. Specifically, the expert receives an observation $x_t^S \sim \mathcal{U}_S(\cdot|s_t)$ from the source-POMDP, whereas the agent observes $x_t^T \sim \mathcal{U}_T(\cdot|s_t)$ from the target-POMDP. Since these observations may not be identical (i.e., $x_t^S \neq x_t^T$), we refer to this discrepancy as \textit{visual mismatch}. This formulation captures scenarios where the agent must infer the expert’s behavior despite differences in visual perception, a common challenge in real-world tasks such as sim-to-real transfer.

\paragraph{Reinforcement learning}  
Aligning with previous work, we define RL in the fully observable Markov Decision Process (MDP) which is a special case of a POMDP where the underlying state $s$ is directly observable, i.e., $\mathcal{X} = \mathcal{S}$ and $\mathcal{U}(s) = \delta_s$, where $\delta_s$ denotes a Dirac distribution centered at $s$.
Given an MDP and a stationary policy $\pi: \mathcal{S} \to P(\mathcal{A})$, the objective is to maximize the expected total discounted return, defined as $J(\pi) = \mathbb{E}_{\tau} [ \sum_{t=0}^{\infty} \gamma^t \mathcal{R}(s_t, a_t)]$, where the trajectory $\tau = (s_0, a_0, s_1, a_1, \dots)$ is generated by following policy $\pi$. A stationary policy $\pi$ induces a normalized discounted state visitation distribution given by $d_{\pi}(s) = (1-\gamma) \sum_{t=0}^{\infty} \gamma^t \mathbb{P}(s_t = s \mid \rho_0, \pi, \mathcal{T})$, representing the expected frequency of visiting state $s$ under $\pi$. The corresponding normalized discounted state-action visitation distribution is $\rho_{\pi}(s, a) = d_{\pi}(s) \pi(a | s)$, which quantifies the expected frequency of encountering state-action pairs $(s, a)$. Furthermore, we define the state-action value function (Q-function) as $Q^{\pi}(s, a) = \mathbb{E}_{\tau} [ \sum_{t=0}^{\infty} \gamma^t \mathcal{R}(s_t, a_t) \mid S_0 = s, A_0 = a]$. Finally, when a policy is parameterized by $\bm{\theta} \in \varTheta \subset \mathbb{R}^k$, we denote it as $\pi_{\bm{\theta}}$.

\paragraph{Generative adversarial imitation learning}\label{sec:prel_gail} Assume we have a set of expert demonstrations $\tau_E = (s_{0:T}, a_{0:T})$ generated by the expert policy $\pi_E$, a set of trajectories $\tau_{\bm{\theta}}$ generated by the policy $\pi_{\bm{\theta}}$, and a discriminator network $D_{\bm{\chi}}: \mathcal{S}\times\mathcal{A} \to [0,1]$ parameterized by $\bm{\chi}$. Generative adversarial IL \cite{ho2016generative} optimizes the min-max objective 
\begin{align}
\begin{split}
    \min_{\bm{\theta}} \max_{\bm{\chi}} \ &\mathbb{E}_{\tau_E}[\log(D_{\bm{\chi}}(s,a))] \\ &+ \mathbb{E}_{\tau_{\bm{\theta}}}[\log(1 - D_{\bm{\chi}}(s,a))]. \label{eq:IRL_disc}
\end{split}
\end{align}
Maximizing \eqref{eq:IRL_disc} with respect to $\bm{\chi}$ is effectively an inverse RL step where a reward function, $r_{\bm{\chi}}(s,a) = -\log(1-D_{\bm{\chi}}(s,a))$, is inferred by leveraging $\tau_E$ and $\tau_{\bm{\theta}}$. Minimizing \eqref{eq:IRL_disc} with respect to $\bm{\theta}$ is an RL step, where the agent aims to minimize its expected cost. Optimizing \eqref{eq:IRL_disc} is equivalent to minimizing $\mathbb{D}_{\text{JS}}(\rho_{\pi_{\bm{\theta}}}(s,a)||\rho_{\pi_E}(s,a))$~\cite{ghasemipour2020divergence}. 

\begin{figure*}
    \centering
    \includegraphics[width=0.8\linewidth]{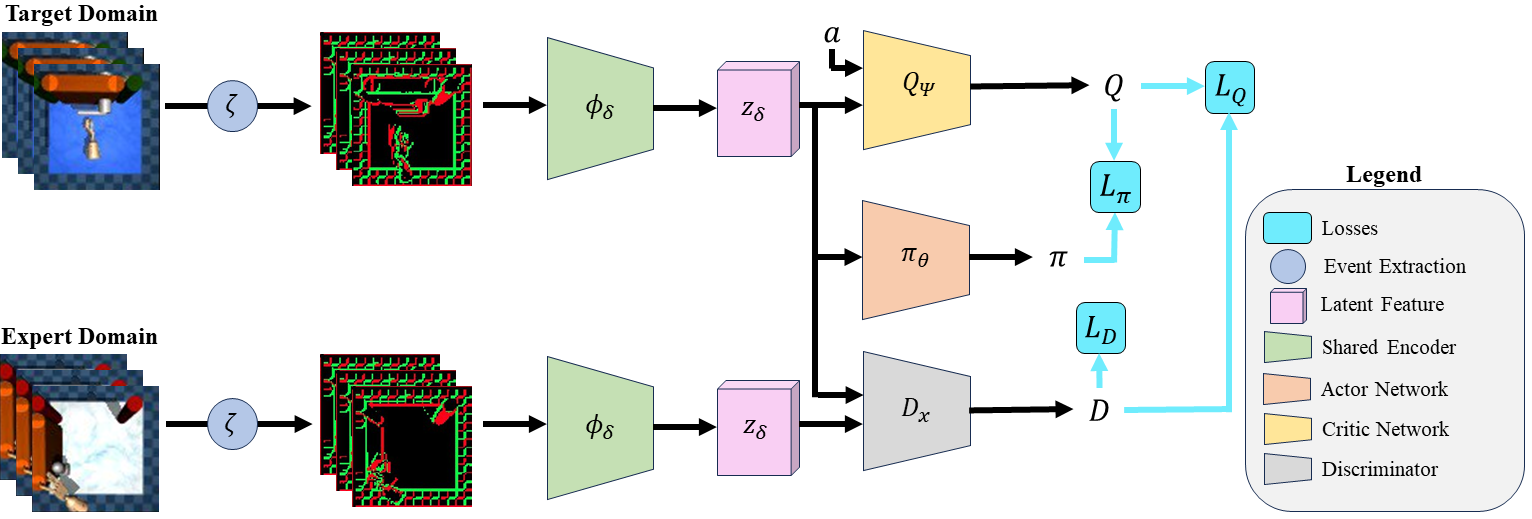}
    \caption{Summary of EB-LAIfO. Given the RGB sequence, the corresponding events stream is extracted following \cref{eq:event_transf}. A feature extractor network $\phi_{\delta}$ is used to generate the latent features $z_\delta$ used by both the Q-function $Q_{\bm{\psi}}$ and discriminator $D_\chi$ for the imitation problem. The discriminator $D_\chi$ is trained as in \eqref{eq:AIL_BCE} and returns the reward function $r_{\bm{\chi}}$ which is then maximized through an RL step. The RL step is described in \eqref{eq:DDPG_critic} and follows the Deep Deterministic Policy Gradient (DDPG) pipeline~\cite{silver2014deterministic}. Our feature extractor network $\phi_{\delta}$ is trained jointly with the Q-function and extracts goal-relevant information directly from the events stream.}
    \label{fig:EB-LAIfO_scheme}
\end{figure*}

\section{EVENT-BASED PERCEPTION FOR IMITATION WITH VISUAL MISMATCH}
Given a target-POMDP and a source-POMDP (i.e. expert), we can categorize the observation space $\mathcal{X}$ into two distinct components: $(i)$ \textit{goal-relevant information}, necessary for successful task completion, and $(ii)$ \textit{visual distractors}, which do not contribute to task execution. Following \cite{giammarino2024visually}, we define $\mathcal{X}$ as $\mathcal{X} = (\bar{\mathcal{X}}, \hat{\mathcal{X}})$, where $\bar{\mathcal{X}}$ represents goal-relevant information that remains consistent across the source-POMDP and target-POMDP, while $\hat{\mathcal{X}}$ denotes visual distractors that vary across domains. Accordingly, the observations in the source and target domains can be expressed as $x_t^S = (\bar{x}_t, \hat{x}_t^S)$ and $x_t^T = (\bar{x}_t, \hat{x}_t^T)$. The objective of our approach is to remove the distracting information ($\hat{x}_t^T,\hat{x}_t^S \in \hat{\mathcal{X}}$) while preserving the goal-relevant information ($\bar{x}_t \in \bar{\mathcal{X}}$). We summarize our solution in the following paragraphs and provide a schematic visualization of the full pipeline in Fig.~\ref{fig:EB-LAIfO_scheme}.

\paragraph{Event-based image transformation} Departing from previous work which enforces domain invariance in a learned feature space $\mathcal{Z}$~\cite{stadie2017third, cetin2020domain, giammarino2024visually}, we propose to extract goal-relevant information $\bar{\mathcal{X}}$ directly from the observation space $\mathcal{X}$. This is accomplished by introducing a transformation $\zeta: \mathcal{X}^2 \to \mathcal{\bar{X}}$, that takes two consecutive RGB frames as input and returns the corresponding event representation as output.
Specifically, in $\bar{x}_t$, each pixel independently encodes an event $E_i = (u, v, t_i, p_i)$ (or the lack of it) when the brightness change exceeds a predefined threshold. Here, $(u,v)$ are the pixel indices, $t_i$ is the time step, and $p_i \in \{+1, -1\}$ represents the polarity of intensity change. An event is triggered when:
\begin{equation*}
    \left| L_t(u,v) - L_{t-1}(u,v) \right| \geq C,
\end{equation*}
where $L_t(u,v) = \log(I_t(u,v))$ represents the logarithm of the pixel intensity $I(u,v)$ at time $t$ and $C$ is a predefined threshold. 
As a result, we obtain $\bar{x}_t = \zeta(x_t, x_{t-1})$ as
\begin{equation}
    \bar{x}_t(u, v) = \begin{cases}
        +1 \ &\text{if} \ L_t(u, v) - L_{t-1}(u, v) \geq C, \\
        -1 \ &\text{if} \ L_t(u, v) - L_{t-1}(u, v) \leq - C, \\
        0 \ &\text{otherwise}.
    \end{cases}
    \label{eq:event_transf}
\end{equation}
Note that in order to avoid fully discarding the whole static content when the camera view is static, we simulate a camera shift by padding $x_t$ by one pixel, i.e., $\bar{x}_t(u,v) = \zeta (x_t(u,v), x_{t-1}(u+1,v+1))$. Importantly, the event transformation in \eqref{eq:event_transf} is invariant across multiple manifolds, including brightness and low-frequency image details, which often contain significant visual distractors. In the following, we formally show this property for both \textit{affine transformation} and \textit{low-frequency transformation}. 

Consider the brightness variation described by the affine transformation $I' = \alpha I + \beta$. By plugging $I'$ in \eqref{eq:event_transf} we obtain
\begin{align*}
    L'_t - L'_{t-1} = 
    \log(\alpha I_t + \beta) -  \log(\alpha I_{t-1} + \beta) = \log\left(\frac{\alpha I_t + \beta}{\alpha I_{t-1} + \beta}\right).
\end{align*}
Assuming $\alpha I >> \beta$ yields
\begin{equation*}
    \approx \log\left(\frac{\alpha I_t }{\alpha I_{t-1} }\right) = \log\left(\frac{I_t }{ I_{t-1} }\right) = \log(I_t)-\log(I_{t-1}) = L_t - L_{t-1}.
\end{equation*}

Furthermore, for low frequency changes we obtain 
\begin{equation*}
\zeta(I_t, I_{t-1} + \eta) = \zeta(I_t, I_{t-1}), \quad \text{if} \quad C - \left| \log \frac{I_t}{I_{t-1}} \right| > \eta,
\end{equation*}
denoting robustness and domain-invariance with respect to $\eta$. It is important to note that such properties are not present in any of the popular image transformations (such as greyscale conversion or edge filtering).

Provided the goal-relevant information $\bar{x}_t$, invariant between the source and target POMDPs, the V-IfO problem with mismatches is effectively reduced to a standard V-IfO problem which can be solved with any SOTA algorithm. Due to its improved computational efficiency, we integrate our approach with LAIfO~\cite{giammarino2023adversarial} and refer to this event-based version as EB-LAIfO. We introduce the full EB-LAIfO pipeline in the following paragraph.

\begin{figure}
    \centering
    \begin{subfigure}{\linewidth}
        \centering
        \includegraphics[width=\linewidth]{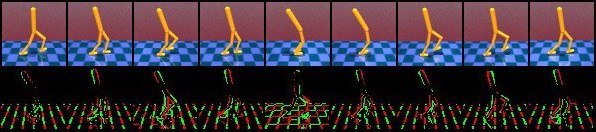}
        \caption{Expert domain}
        \label{fig:pitotino_transform_a}
    \end{subfigure}
    
    
    \begin{subfigure}{\linewidth}
        \centering
        \includegraphics[width=\linewidth]{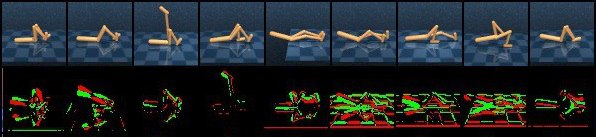}
        \caption{Target Domain}
        \label{fig:pitotino_transform_b}
    \end{subfigure}
    
    \caption{Examples scenes from the expert and target domain, both in original RGB space and obtained event spaced (green is positive event, red is negative).}
    \label{fig:pitotino_transform}
\end{figure}

\paragraph{Event-based latent adversarial imitation from observations}  
Suppose we have access to a video of an expert demonstrating a task, denoted as $\tau_E = x^S_{0:T}$. Our goal is to perform AIL, as introduced in Section~\ref{sec:prel_gail}, directly from this video. To address this challenge, we first estimate a latent variable $z \in \mathcal{Z}$ from the observation space and then execute the entire AIL pipeline within this latent space. However, naively applying this approach in our setting results in poor performance due to the inherent visual mismatch between the source and target domains, coupled with the absence of expert action data in the dataset. These challenges make latent variable estimation significantly more difficult~\cite{giammarino2023adversarial, giammarino2024visually} (see also Section 6.10 in~\cite{giammarino2024use}). To mitigate these issues, we first apply our event-based transformation, introduced in \eqref{eq:event_transf}, to process all observation streams collected from both the source and target POMDPs. Then, within our event-based observation space $\bar{\mathcal{X}}_{\text{EB}}$, we define a feature extractor $\phi_{\bm{\delta}}:\bar{\mathcal{X}}_{\text{EB}}^d\to\mathcal{Z}$, which takes as input a stack of $d \in \mathbb{N}$ observations such that $z = \phi_{\bm{\delta}}(\bar{x}_{t^-:t})$, where $t - t^- + 1 = d$. The latent variable $z \in \mathcal{Z}$ is estimated directly in $\bar{\mathcal{X}}_{\text{EB}}$ by training the feature extractor $\phi_{\bm{\delta}}$ jointly with the critic networks $Q_{\bm{\psi}_k}$ ($k=1,2$). Specifically, we minimize the following loss
\begin{align}
    \begin{split}
        \mathcal{L}_{\bm{\delta}, \bm{\psi}_k}(\mathcal{B}) &= \mathbb{E}_{(z, a, z')\sim\mathcal{B}}[(Q_{\bm{\psi}_k}(z, a) - y)^2], \\
        y &= r_{\bm{\chi}}(z,z') + \gamma \min_{k={1,2}} Q_{\bar{\bm{\psi}}_k}(z', a'),  
    \end{split} \label{eq:DDPG_critic} 
\end{align}
where $a$ is an action stored in the agent buffer $\mathcal{B}$ and which is used by the agent to interact with the environment, while $a' = \pi_{\bm{\theta}}(z') + \epsilon$ where $\epsilon \sim \text{clip}(\mathcal{N}(0,\sigma^2), -c, c)$ is a clipped exploration noise with $c$ the clipping parameter and $\mathcal{N}(0,\sigma^2)$ a univariate normal distribution with zero mean and $\sigma$ standard deviation. Finally, $\bar{\bm{\psi}}_1$ and $\bar{\bm{\psi}}_2$ are the slow-moving weights for the target Q networks. 

The reward function $r_{\bm{\chi}}(z, z')$ in \eqref{eq:DDPG_critic} is defined as $r_{\bm{\chi}}(z, z') = - \log\big(1 - D_{\bm{\chi}}(z, z')\big)$, where $D_{\bm{\chi}}:\mathcal{Z}\times\mathcal{Z} \to [0,1]$ is a discriminator function trained to optimize the following loss, as in \eqref{eq:IRL_disc}:  
\begin{align}
    \begin{split}
        \max_{\bm{\chi}} \ \  &\mathbb{E}_{(z,z') \sim \mathcal{B}_E(\tau_E)}[\log(D_{\bm{\chi}}(z,z'))] \\
        &+ \mathbb{E}_{(z,z') \sim \mathcal{B}(\tau_{\bm{\theta}})}[\log(1 - D_{\bm{\chi}}(z,z'))].
    \end{split}\label{eq:AIL_BCE} 
\end{align}  
Here, $\mathcal{B}_E$ and $\mathcal{B}$ are the replay buffers storing $\tau_E$ and $\tau_{\bm{\theta}}$, corresponding to the expert and learning agent observations, respectively. In this final step, both expert and learning agent observations are first preprocessed using our event-based transformation such that  
$\bar{x}_t = \zeta(x_t^S, x_{t-1}^S)$ and $\bar{x}_t = \zeta(x_t^T, x_{t-1}^T)$ and mapped onto $\mathcal{X}_{\text{EB}}$. Furthermore, they are embedded as $z = \phi_{\bm{\delta}}(\bar{x}_{t^-:t})$ to formulate the problem compactly in $\mathcal{Z}$. To streamline notation, we write $(z,z') \sim \mathcal{B}(\tau_{\bm{\theta}})$ and $(z,z') \sim \mathcal{B}_E(\tau_E)$. As mentioned, the entire pipeline for EB-LAIfO is summarized in Fig.~\ref{fig:EB-LAIfO_scheme}.

\begin{table*}[ht!]
\centering
\caption{Summary of the experiments on the DeepMind control suite~\cite{tassa2018deepmind}. We train all the algorithms for 500,000 steps. The learned policies are evaluated based on average return over $10$ episodes. We report the mean and standard deviation of the final return across $5$ seeds and \textbf{highlight} the best performance. These results highlight the effectiveness of EB-LAIfO in handling light and color mismatches compared to the tested baselines.}
\label{table_dmc}
    \begin{tabular}{c | c c c c c}
        \toprule
        \multicolumn{6}{c}{\makecell{Target Env: \raisebox{-.5\height}{\includegraphics[width=0.08\linewidth]{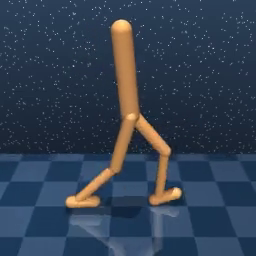}}, Expert Performance = $950$}} \\
        \cmidrule(lr){1-6}
         & Light & Body & Floor & Background & Full \\
        Source Env & \raisebox{-.5\height}{\includegraphics[width=0.08\linewidth]{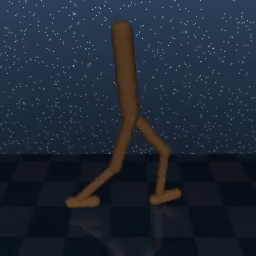}} & \raisebox{-.5\height}{\includegraphics[width=0.08\linewidth]{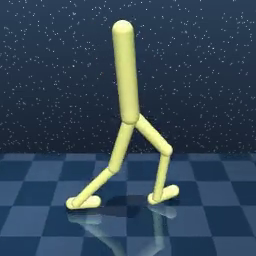}} & \raisebox{-.5\height}{\includegraphics[width=0.08\linewidth]{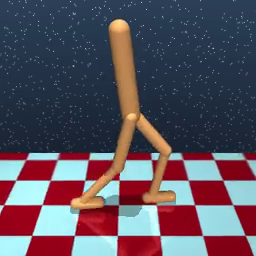}} & \raisebox{-.5\height}{\includegraphics[width=0.08\linewidth]{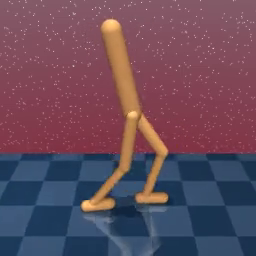}} & \raisebox{-.5\height}{\includegraphics[width=0.08\linewidth]{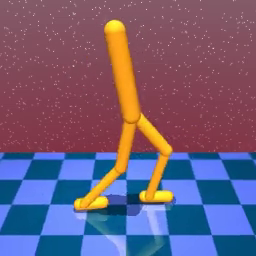}} \\
        \cmidrule(lr){1-6}
        EB-LAIfO (ours) & $\bm{677 \pm 285}$ & $\bm{854 \pm 125}$ & $\bm{603 \pm 271}$ & $\bm{858 \pm 108}$ & $\bm{898 \pm 48}$ \\
        C-LAIfO~\cite{giammarino2024visually} & $202 \pm 287$ & $778 \pm 155$ & $123 \pm 55$ & $441 \pm 328$ & $203 \pm 78$ \\
        LAIfO~\cite{giammarino2023adversarial} w/ data aug & $63 \pm 28$ & $741 \pm 102$ & $353 \pm 200$ & $69 \pm 48$ & $142 \pm 230$ \\
        DisentanGAIL~\cite{cetin2020domain} & $26 \pm 4$ & $425 \pm 247$ & $46 \pm 16$ & $25 \pm 5$ & $26 \pm 6$ \\
        PatchAIL~\cite{liu2022visual} w/ data aug & $22 \pm 8$ & $170 \pm 65$ & $14 \pm 1$ & $30 \pm 6$ & $40 \pm 19$ \\
        \bottomrule
    \end{tabular}
\end{table*}

\section{EXPERIMENTS}
\label{sec:experiments}
In this section, we first demonstrate how EB-LAIfO effectively handles various types of visual mismatches in the V-IfO setting (Sec.~\ref{sec:experiments_V-IfO}); and then showcase how EB-LAIfO is more effective than other algorithms at facilitating learning in challenging robotic manipulation tasks with sparse rewards and realistic visual inputs (Sec.~\ref{sec:experiments_RL+IL}). Finally, in Sec.~\ref{sec:ablation}, we study the robustness of our approach to noise in the event stream.
Unless specified, we use DDPG~\cite{lillicrap2015continuous} to train experts in a fully observable setting and collect $100$ episodes of expert videos for imitation.

\subsection{Visual Imitation from Observations with mismatch}
\label{sec:experiments_V-IfO}
\begin{figure}
    \centering
    \includegraphics[width=\linewidth]{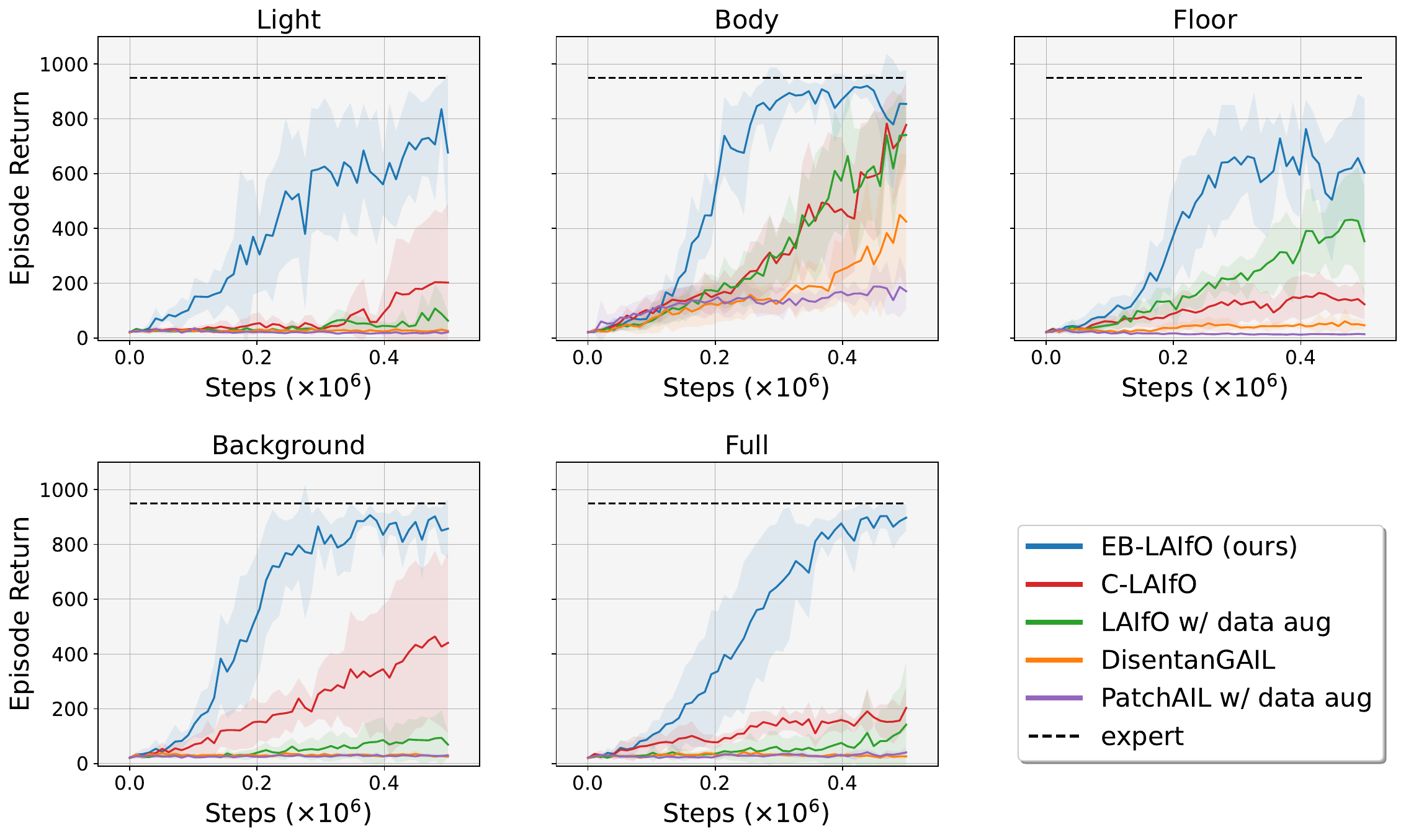}
    \caption{Learning curves for the results in Table~\ref{table_dmc}. Plots show the average return per episode and the standard deviation across seeds as a function of training steps.}
    \label{fig:dmc_curves}
\end{figure}

In this paragraph, we evaluate EB-LAIfO in the V-IfO setting across various visual mismatches and compare it with four baselines: C-LAIfO~\cite{giammarino2024visually}, LAIfO~\cite{giammarino2023adversarial}, PatchAIL~\cite{liu2022visual}, and DisentanGAIL~\cite{cetin2020domain}. For the Light experiment, we provide our baselines with brightness transformation as a data augmentation technique, while for the other experiments, we define data augmentation as a color transformation. The results, summarized in Table~\ref{table_dmc}, are further illustrated through learning curves in Fig.~\ref{fig:dmc_curves}. In our assessment of performance, we consider both final results and learning efficiency.  

The results in Table~\ref{table_dmc} demonstrate that EB-LAIfO consistently outperforms all other baselines in handling visual mismatches. Notably, EB-LAIfO reaches near-expert performance in the Full setting (\(898 \pm 48\)), achieving a fourfold improvement over C-LAIfO (\(203 \pm 78\)). Furthermore, when the discrepancy in final performance is smaller, such as in the Body setting, we observe that EB-LAIfO converges significantly more efficiently than all other tested methods.  

The key reasons for this superior performance are twofold. First, event-based perception retains goal-relevant information more effectively than data augmentation-based techniques, which heavily depend on the types of augmentations and are more susceptible to stochasticity, as augmentations are randomized. Second, since our event-based transformation is a purely mathematical operation that does not require active learning, our approach is inherently more efficient than learning-based methods. For example, the contrastive learning step in C-LAIfO requires extensive training and data augmentation to extract domain-invariant features, whereas event-based transformations enhance this invariance by filtering out visual distractors directly from the pixel space.  

This advantage leads to improved performance and sample efficiency, particularly in environments with strong visual mismatches, such as Background (\(858 \pm 108\) for EB-LAIfO vs. \(441 \pm 328\) for C-LAIfO) and Floor (\(603 \pm 271\) for EB-LAIfO vs. \(123 \pm 55\) for C-LAIfO). Furthermore, the limitations of traditional methods are evident, as DisentanGAIL, LAIfO and PatchAIL fail to handle domain shifts effectively, achieving low scores for most of the mismatches.  

These results confirm that EB-LAIfO's event-based perception framework is significantly more robust and sample-efficient than other learning-based approaches, making it a promising solution for real-world IL in visually mismatched environments.

\subsection{Dexterous manipulation experiments} 
\label{sec:experiments_RL+IL}

\begin{table}
\centering
\scriptsize
\caption{Summary of the experiments on the Adroit platform for dynamic dexterous manipulation~\cite{Kumar2016thesis}. The Door-Light and Door-Color experiments consider \eqref{fig:door} as the source-POMDP and \eqref{fig:door-light} and \eqref{fig:door-color} as the respective target-POMDPs. Similarly, the Hammer-Light and Hammer-Color experiments consider \eqref{fig:hammer} as source-POMDP and \eqref{fig:hammer-light} and \eqref{fig:hammer-color} as the respective target-POMDPs. We use the VRL3 in \cite{wang2022vrl3} to train expert policies and collect $100$ episodes of expert data. All the algorithms are trained for $10^6$ steps. The learned policies are evaluated based on average return over $10$ episodes. We report the mean and standard deviation of the final return across $4$ seeds and \textbf{highlight} the best performance. }
\label{table_adroit}
    \begin{tabular}{c | c c | c c }
        \toprule
        & \multicolumn{2}{c|}{Door} & \multicolumn{2}{c}{Hammer} \\
        & Light & Color & Light & Color \\
        \cmidrule(lr){1-5}
        Expert & \multicolumn{2}{c|}{$170$} & \multicolumn{2}{c}{$184$} \\
        \cmidrule(lr){1-5}
        RL+EB-LAIfO (ours) & $111 \pm 59$ & \bm{$106 \pm 48$} & \bm{$174 \pm 10$} & \bm{$177 \pm 9$} \\
        RL+C-LAIfO~\cite{giammarino2024visually} & \bm{$150 \pm 5$} & $79 \pm 81$ & $99 \pm 76$ & $118 \pm 74$ \\
        RL+LAIfO~\cite{giammarino2023adversarial} & $35 \pm 64$ & $68 \pm 71$ & $82 \pm 85$ & $-2 \pm 0.0$ \\
        \bottomrule
    \end{tabular}
\end{table}

\begin{figure}
    \centering
    \includegraphics[width=0.8\linewidth]{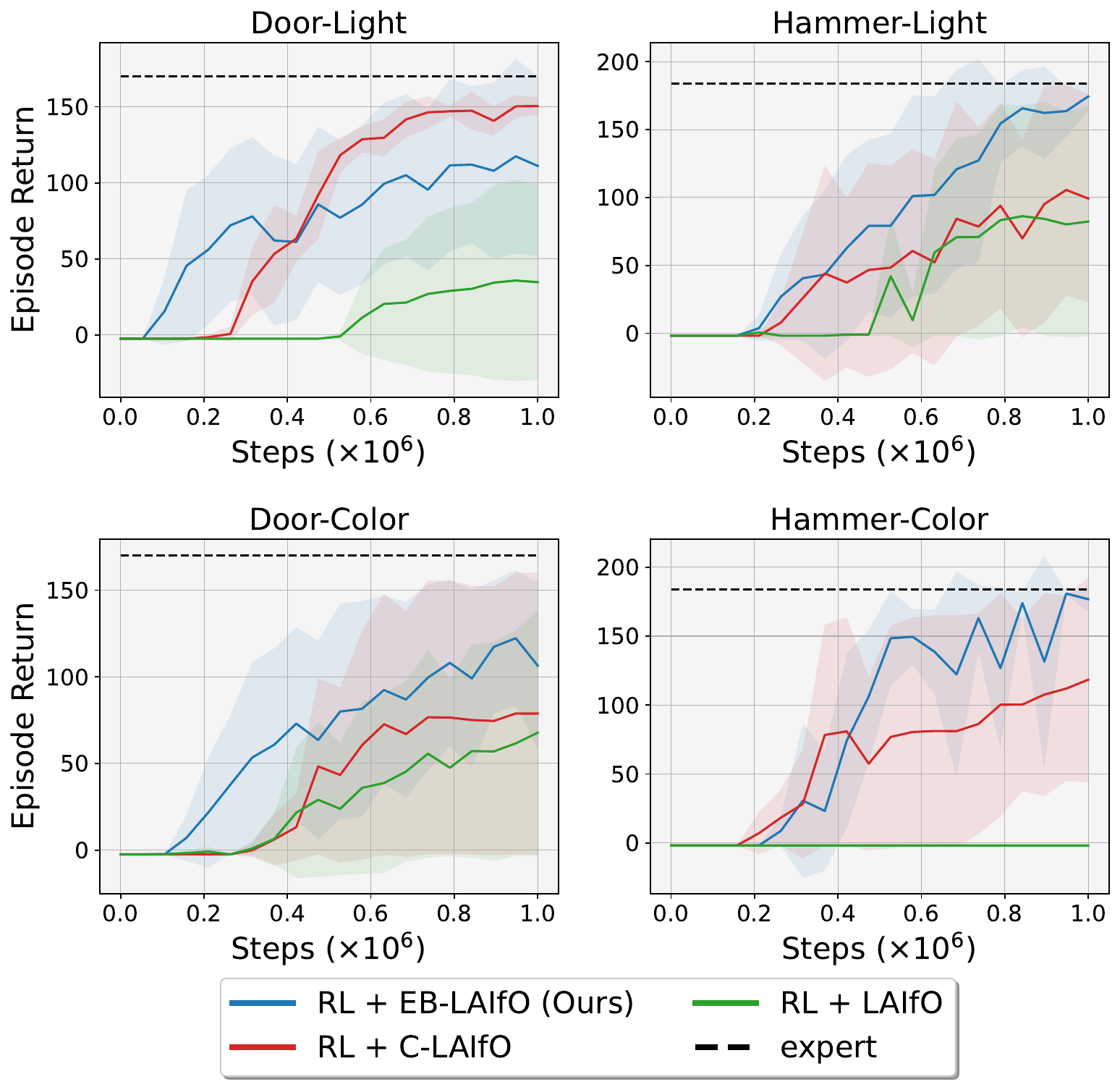}
    \caption{Learning curves for the results described in Table~\ref{table_adroit}. Plots show the average return per episode and the standard deviation across seeds as a function of training steps.}
    \label{fig:learning_curves_adroit}
\end{figure}

\begin{figure}[ht!]
    \centering
    \begin{subfigure}[t]{0.15\linewidth}
        \centering
        \includegraphics[width=\linewidth]{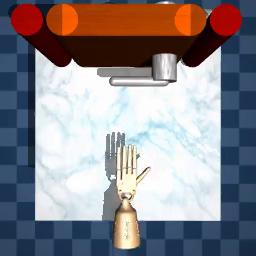}
        \caption{}
        \label{fig:door}
    \end{subfigure}
    \begin{subfigure}[t]{0.15\linewidth}
        \centering
        \includegraphics[width=\linewidth]{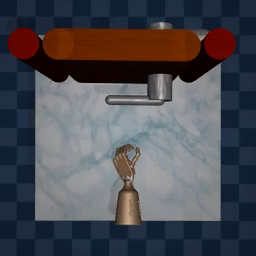}
        \caption{}
        \label{fig:door-light}
    \end{subfigure}
    \begin{subfigure}[t]{0.15\linewidth}
        \centering
        \includegraphics[width=\linewidth]{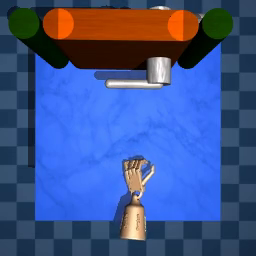}
        \caption{}
        \label{fig:door-color}
    \end{subfigure}
    \begin{subfigure}[t]{0.15\linewidth}
        \centering
        \includegraphics[width=\linewidth]{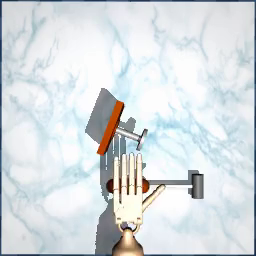}
        \caption{}
        \label{fig:hammer}
    \end{subfigure}
    \begin{subfigure}[t]{0.15\linewidth}
        \centering
        \includegraphics[width=\linewidth]{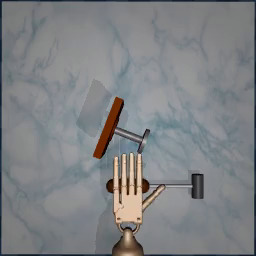}
        \caption{}
        \label{fig:hammer-light}
    \end{subfigure}
    \begin{subfigure}[t]{0.15\linewidth}
        \centering
        \includegraphics[width=\linewidth]{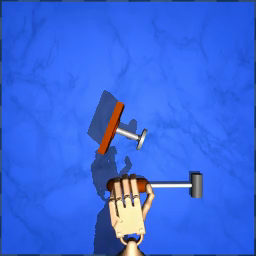}
        \caption{}
        \label{fig:hammer-color}
    \end{subfigure}
    \caption{Adroit environments used for the experiments in Table~\ref{table_adroit}.}
    \label{fig:adroit}
\end{figure}

\begin{figure}[ht!]
    \centering
    \begin{subfigure}[t]{0.2\linewidth}
        \centering
        \includegraphics[width=\linewidth, height=1.55cm]{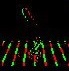}
        \caption{$\sigma = 0.01$}
        \label{fig:05}
    \end{subfigure}
    ~
    \begin{subfigure}[t]{0.2\linewidth}
        \centering
        \includegraphics[width=\linewidth, height=1.55cm]{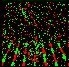}
        \caption{$\sigma = 0.1$}
        \label{fig:1}
    \end{subfigure}
    ~
    \begin{subfigure}[t]{0.2\linewidth}
        \centering
        \includegraphics[width=\linewidth, height=1.55cm]{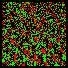}
        \caption{$\sigma = 0.2$}
        \label{fig:2}
    \end{subfigure}
    ~
    \begin{subfigure}[t]{0.2\linewidth}
        \centering
        \includegraphics[width=\linewidth, height=1.55cm]{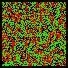}
        \caption{$\sigma = 0.5$}
        \label{fig:5}
    \end{subfigure}
    \caption{Visualization of the resulting event stream with varying noise standard deviation level $\sigma$. }
    \label{fig:noise}
\end{figure}

In this section, we evaluate our algorithm on a series of challenging robotic manipulation tasks from the Adroit platform for dynamic dexterous manipulation~\cite{Kumar2016thesis}. These experiments demonstrate how the reward \( r_{\bm{\chi}} \), learned by EB-LAIfO from expert videos, can be effectively combined with a sparse reward \( \mathcal{R} \), collected by the agent through interaction with the environment, to enhance learning efficiency. The RL problem aims to maximize the total reward 
\begin{equation}
    \mathcal{R}_{\text{tot}} = \mathcal{R}(s_t,a_t) + r_{\bm{\chi}}(z_t, z_{t+1}),
    \label{eq:R_tot}
\end{equation}
where $r_{\bm{\chi}}(z, z') = - \log\big(1 - D_{\bm{\chi}}(z, z')\big)$ with $D_{\bm{\chi}}(z, z')$ trained to optimize \eqref{eq:AIL_BCE}. This approach is particularly relevant for robotic tasks, where sparse rewards are often the most feasible option in real-world settings. However, relying solely on sparse rewards can make learning challenging and inefficient. In this context, leveraging expert videos can significantly enhance efficiency. 

To evaluate the effectiveness of EB-LAIfO, we compare it against C-LAIfO and the standard LAIfO algorithm on the Adroit platform for dynamic dexterous manipulation~\cite{Kumar2016thesis}. All these methods employ an encoder to process pixel observations, extracting embeddings in $\mathcal{Z}$, which are then concatenated with robot sensory observations. Notably, the expert’s sensory observations are not used in the imitation process, as we assume access only to expert videos. Consequently, in these experiments, we seek to maximize \( \mathcal{R}_{\text{tot}} \) in \eqref{eq:R_tot}, rather than just \( r_{\bm{\chi}} \), as in standard imitation learning.  

The results, summarized in Table~\ref{table_adroit}, highlight the advantages of EB-LAIfO in handling visual mismatches. Across all conditions, EB-LAIfO consistently outperforms LAIfO and exhibits strong performance compared to C-LAIfO. In the Hammer tasks, EB-LAIfO achieves near-expert performance with \(174 \pm 10\) in the Light setting and \(177 \pm 9\) in the Color setting, demonstrating its robustness in retaining task-relevant features despite visual perturbations. In contrast, C-LAIfO shows reduced performance in these settings (\(99 \pm 76\) and \(118 \pm 74\), respectively), indicating that it struggles to fully bridge domain gaps introduced by color variations.  

In the Door tasks, C-LAIfO achieves the best performance in the Light setting (\(150 \pm 5\)), slightly surpassing EB-LAIfO (\(111 \pm 59\)), suggesting that C-LAIfO may be more effective in handling brightness-based domain shifts in this scenario. However, EB-LAIfO significantly outperforms C-LAIfO in the more challenging Door-Color setting (\(106 \pm 48\) vs. \(79 \pm 81\)), reinforcing the idea that event-based transformations are particularly effective at mitigating challenging color-based domain mismatches. LAIfO exhibits the weakest performance across all experiments, failing to generalize effectively under domain shifts. In the Hammer-Color setting, it even fails to learn a meaningful policy, achieving a negative return (\(-2 \pm 0.0\)), while in the Door-Light setting, it only reaches \(35 \pm 64\), far below the other methods.  

These results confirm that EB-LAIfO provides a robust and sample-efficient approach for imitation in visually mismatched environments, particularly when handling complex domain shifts such as those introduced by color variations.

\subsection{Event noise ablation}
\label{sec:ablation}
\begin{figure}
    \centering
    \includegraphics[width=0.9\linewidth]{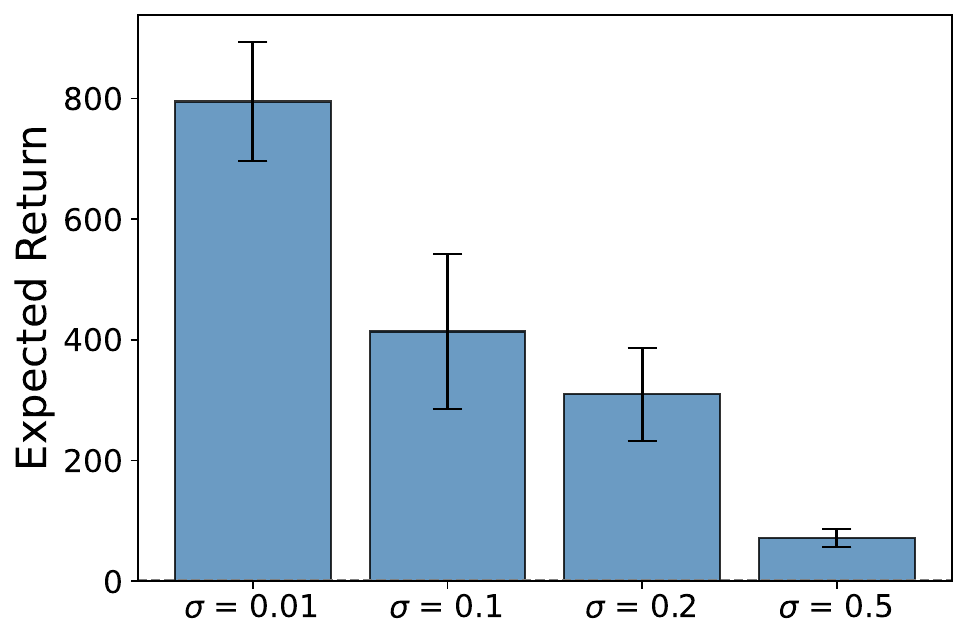}
    \caption{Ablation study with different $\sigma$ noise values. Each policy is evaluated as in Table~\ref{table_dmc}. We report mean and standard deviation across the last $100$ evaluations for each run.}
    \label{fig:ablation_numbers}
\end{figure}
To assess the robustness of our approach, as well as proving its applicability for real event cameras, we perform an ablation study injecting Gaussian noise to simulate real-world sensor imperfections and environmental interference. In these experiments, our transformation function $\zeta$ becomes
\begin{equation*}
    \bar{x}_t(u, v) = \begin{cases}
        +1 \ &\text{if} \ L_t(u, v) - L_{t-1}(u, v) + \mathcal{N}(0, \sigma^2) \geq C, \\
        -1 \ &\text{if} \ L_t(u, v) - L_{t-1}(u, v) + \mathcal{N}(0, \sigma^2) \leq - C, \\
        0 \ &\text{otherwise},
    \end{cases}
\end{equation*}
where $\sigma$ represents the (fixed) standard deviation of the additive Gaussian noise. While there are other types of noise occurring in real-world scenarios, Gaussian noise provides a reasonable approximation of the combined effect of multiple noise sources, making it an appropriate choice for our robustness evaluation.

We consider four noise levels with $\sigma_{1,..,4}$ set respectively to $0.01,0.1,0.2$ and $0.5$. Resulting event samples are visualized in Fig.~\ref{fig:noise}. The final results, illustrated in Fig~\ref{fig:ablation_numbers}, reveal that our approach maintains robust performance up to moderate noise levels ($\sigma=0.2$), with outcomes comparable to the noise-free baseline. While performance degradation becomes evident at higher noise intensities, it's worth noting that such extreme noise levels (as illustrated in Fig.~\ref{fig:noise}) severely corrupt the sensor information to an extent that would be unrealistic for properly functioning event cameras in practical applications.
\section{Conclusion}
\label{sec:conclusion}

In this paper, we introduce a novel approach to V-IfO that addresses the fundamental challenge of visual domain gaps between demonstration and deployment environments. 
By drawing inspiration from biological vision systems and event camera technology, we developed a lightweight, event-inspired perception module that inherently filters out appearance-based distractors while preserving the motion dynamics critical for successful imitation.
Our experiments demonstrate that discretized temporal gradients provide a robust alternative to conventional learning-based methods. This approach eliminates the need for expensive domain randomization or handcrafted data augmentation strategies that have limited current SOTA methods.


A limitation of our approach arises in scenarios where appearance features, rather than just motion dynamics, provide critical contextual cues for task completion. Another limitation of our current implementation lies in the fidelity of our synthetic event stream generation from RGB videos. Our model does not fully capture all secondary effects of real-world event camera systems, such as complex noise models and asynchronous data streams. While sufficient as a proof-of-concept, real event cameras capture temporal information at microsecond resolution, which our conversion process cannot fully replicate. An interesting direction for future work is to apply our method in real-world applications, integrating either standard RGB or event cameras. Additionally, we believe our approach offers a promising alternative for bridging another common domain gap, namely the one between synthetic and real-world visual data. From an algorithmic perspective, exploring different representation learning techniques for extracting goal-relevant features from event-based images represents another exciting avenue. Methods that better exploit the unique characteristics of event-based data streams could further enhance performance and improve efficiency, not only for IL but also for RL in POMDP settings.


\bibliography{mybib}


\end{document}